\documentclass{amia}
\usepackage{lipsum} 
\usepackage{hyperref}       
\usepackage{url}            
\usepackage{booktabs}       
\usepackage{amsfonts}       
\usepackage{enumitem}
\usepackage{amsmath}
\usepackage{amssymb}
\usepackage{multirow}
\usepackage{caption}
\usepackage{subcaption}
\usepackage{bm}

\usepackage{listings}
\usepackage{xcolor}
\usepackage{verbatim}
\usepackage{lipsum}

\lstdefinestyle{mystyle}{
    basicstyle=\ttfamily,
    numbers=none,
    breaklines=true,
    backgroundcolor=\color{gray!10},
    frame=single,
    captionpos=b,
    xleftmargin=1.5em,
    xrightmargin=1.5em,
    linewidth=\textwidth
}

\begin{document}

\title{Can LLMs Support Medical Knowledge Imputation? An Evaluation-Based Perspective}

\author{Xinyu Yao, MSc$^1$, Aditya Sannabhadti, BSc$^1$ , Holly Wiberg, PhD$^1$, Karmel S. Shehadeh, PhD$^2$, Rema Padman, PhD$^1$}

\institutes{
    $^1$Heinz College of Information Systems and Public Policy, Carnegie Mellon University, Pittsburgh, PA\\  $^2$Daniel J. Epstein Department of Industrial and Systems Engineering, University of Southern California, Los Angeles, CA
}

\maketitle

\section*{Abstract}
\textit{Medical knowledge graphs (KGs) are essential for clinical decision support and biomedical research, yet they often exhibit incompleteness due to knowledge gaps and structural limitations in medical coding systems. This issue is particularly evident in treatment mapping, where coding systems such as ICD, Mondo, and ATC lack comprehensive coverage, resulting in missing or inconsistent associations between diseases and their potential treatments. To address this issue, we have explored the use of Large Language Models (LLMs) for imputing missing treatment relationships. Although LLMs offer promising capabilities in knowledge augmentation, their application in medical knowledge imputation presents significant risks, including factual inaccuracies, hallucinated associations, and instability between and within LLMs. In this study, we systematically evaluate LLM-driven treatment mapping, assessing its reliability through benchmark comparisons. Our findings highlight critical limitations, including inconsistencies with established clinical guidelines and potential risks to patient safety. This study serves as a cautionary guide for researchers and practitioners, underscoring the importance of critical evaluation and hybrid approaches when leveraging LLMs to enhance treatment mappings on medical knowledge graphs.}

\section{Introduction}
\label{sec:intro}

Medical knowledge graphs (KGs) have become indispensable tools in healthcare information systems, offering structured representations of complex medical information and supporting a wide range of applications. By integrating diverse entities such as diseases, treatments, drugs, and genes from different medical ontologies and coding systems, KGs facilitate semantic reasoning and interoperability across various healthcare information systems. When carefully curated, these graphs enable clinical decision support systems, biomedical discovery, and pharmacological analysis by allowing more interpretable and scalable representations of medical knowledge. Their structured nature makes them particularly suitable for downstream tasks such as drug repurposing, adverse event prediction, and personalized treatment planning.

Despite their value, medical KGs are often incomplete due to both structural limitations and knowledge gaps in their source ontologies. This incompleteness is especially pronounced in the context of treatment mapping, that is, linking diseases to their valid therapeutic interventions, where many associations are either sparsely represented, outdated, or entirely missing. Existing coding systems like ICD, MONDO, and ATC frequently lack the granularity or consistency required to fully capture disease-treatment relationships, resulting in fragmented knowledge bases. These gaps reduce the effectiveness of downstream reasoning systems and may contribute to suboptimal clinical decision-making. Bridging these missing links is crucial to improving the comprehensiveness and accuracy of knowledge-based healthcare technologies.

In recent years, the rapid advancement of large language models (LLMs), such as GPT, Perplexity, Gemini, and Claude, has opened up new opportunities to enhance medical KGs through automated knowledge imputation. These models exhibit strong capabilities in biomedical text understanding, knowledge synthesis, and natural language inference, making them appealing tools for identifying missing relationships not explicitly captured in structured ontologies. In the case of treatment mapping, LLMs can potentially draw from clinical literature, drug labels, and publicly available guidelines to generate novel associations between diseases and drugs. By leveraging such unstructured or semi-structured sources, LLMs offer a scalable mechanism to extend existing medical KGs and support clinical reasoning in low-resource or rapidly evolving medical domains.

However, incorporating LLM-generated knowledge into medical KGs introduces significant risks. Unlike curated ontologies that are grounded in clinical evidence and expert validation, LLMs may produce hallucinated outputs—assertions that are syntactically plausible but factually incorrect or clinically irrelevant. These models can also reflect biases from their pretraining data, misinterpret ambiguous terminology, or contradict established medical guidelines. Furthermore, results generated by different LLMs can be inconsistent with one another, and even a single LLM may exhibit internal instability, producing varying outputs across different runs or over time. In clinical applications where the cost of misinformation is high, such limitations can compromise patient safety and erode trust in AI-powered decision support systems. Therefore, rigorous evaluation is essential before integrating LLM-derived information into high-stakes medical infrastructures.

While LLMs offer promising capabilities for augmenting incomplete medical knowledge graphs, evaluating the reliability of their generated outputs remains a critical challenge, particularly in the absence of sufficient expert resources. Manual review by clinical professionals is often prohibitively time-consuming and expensive, making large-scale validation infeasible. To address this constraint, we adopt a scalable, knowledge-grounded evaluation strategy by comparing LLM-generated treatment relationships against curated, ontology-based knowledge graphs. This approach enables us to assess not only whether LLMs can recover known relationships, but also how well their outputs align with clinically accepted standards. By quantifying the overlap and divergence between generated associations and established sources, we gain insight into both the strengths and limitations of LLMs as knowledge imputers. Our findings show that while LLMs can surface novel and potentially useful treatment suggestions, they also produce inconsistencies that underscore the need for hybrid frameworks combining generative models with domain-grounded validation mechanisms.

This work contributes to the growing literature on trustworthy AI in healthcare by examining how generative models can be used to augment medical knowledge while maintaining safety and interpretability. Our proposed evaluation framework offers a pragmatic solution to the expert-scarcity problem and sheds light on the boundary between meaningful augmentation and unsafe hallucination. By systematically analyzing LLM-generated treatment mappings in the context of existing medical KGs, we offer practical guidance for researchers and practitioners seeking to leverage these powerful models responsibly. Ultimately, our study underscores the importance of cautious integration and transparent validation when deploying LLMs to enhance clinical knowledge infrastructures.

\section{Literature}
\label{sec:literature}

\paragraph{Traditional Approaches to Knowledge Graph Completion}

To address graph incompleteness, a wide range of KG completion techniques have been proposed. Early efforts include translational and semantic matching models such as TransE~\cite{bordes2013translating}, DistMult~\cite{yang2014embedding}, ComplEx~\cite{trouillon2016complex}, and RotatE~\cite{sun2019rotate}, which project entities and relations into continuous vector spaces to predict missing links. Later, graph neural network (GNN) architectures like Relational GCN (R-GCN)~\cite{schlichtkrull2018modeling} and CompGCN\cite{vashishth2019composition} have been applied to KGs to exploit graph structure and multi-relational patterns. These models have shown strong performance on benchmark datasets and have been adopted in biomedical contexts, including MedGCN~\cite{mao2022medgcn} and Decagon~\cite{zitnik2018modeling}, which predict treatment effects and drug-drug interactions by leveraging heterogeneous biological and clinical information. However, such models rely on observed graph topology and known triples during training, making it difficult for them to infer entirely novel relationships absent from the original data. This limits their effectiveness in scenarios where knowledge gaps are structural rather than simply incomplete observations.

\paragraph{Large Language Models for Biomedical Knowledge Augmentation}

With the emergence of LLMs, such as GPT-4, BioGPT~\cite{luo2022biogpt}, and Med-PaLM~\cite{singhal2023large}, there is growing interest in their potential for augmenting structured knowledge bases through generative means. These models, trained on large-scale biomedical corpora, have demonstrated proficiency in information extraction, question-answering, and summarization tasks. Several studies have explored prompting LLMs to generate subject–relation–object triples or biomedical facts that could be incorporated into knowledge graphs~\cite{soman2024biomedical,zhang2024knowgpt}. By capturing latent associations from unstructured literature and guidelines, LLMs offer the promise of recovering valid but undocumented relationships—particularly in domains where ontologies lag behind evolving clinical knowledge. However, LLMs are also prone to hallucination, generating statements that are linguistically plausible yet factually incorrect or unverifiable~\cite{marcus2020next}. In high-stakes domains like medicine, where clinical errors have serious consequences, such hallucinations present a major barrier to practical deployment.

\paragraph{Evaluating LLM-Generated Medical Knowledge}

The challenge of validating LLM-generated knowledge has led to a growing body of work on evaluation techniques. While expert annotation remains the gold standard, it is costly and not scalable. Alternative methods, such as self-consistency checking~\cite{manakul2023selfcheckgpt} or retrieval-augmented generation~\cite{lewis2020retrieval}, have been proposed to detect hallucinations or enhance factual grounding. Nevertheless, these methods are typically developed for general-domain tasks and have limited applicability to the nuanced requirements of medical reasoning. Some recent efforts attempt to align generated knowledge with existing databases or ontologies to assess factuality~\cite{ yao2023extracting,hamed2024fact}, though such approaches have not been widely applied in other biomedical contexts.

Despite recent progress, most existing studies applying LLMs to KG completion focus on general-domain or commonsense knowledge, with limited exploration of clinical-grade biomedical graphs. Evaluation strategies are often either heuristic or purely qualitative, lacking structured, domain-specific benchmarks. Our work addresses this challenge by introducing a systematic evaluation framework based on coverage, alignment, and robustness checks, which enables comparison both among LLM-generated treatment triples and against verified relationships in existing medical KGs. This approach offers a scalable and interpretable proxy for expert review, providing a foundation for safe and transparent integration of LLM-generated knowledge into clinical applications.

\section{Data and Methods}
\label{sec:dm}

This study investigates the extent to which LLMs can generate clinically relevant treatment relationships that align with existing medical KGs. Our methodology consists of three main components: (1) preparing a reference KG by integrating existing ontologies and treatment relationship datasets; (2) querying multiple LLMs with diverse prompts to generate disease-treatment associations; and (3) evaluating the alignment between LLM-generated outputs and curated KG relationships to assess factual accuracy and coverage.

\paragraph{Reference Knowledge Graph Construction}

To provide a reference standard for evaluating LLM responses, we constructed a focused subset of a medical knowledge graph that captures disease-treatment relationships. Entities and relationships were extracted from established biomedical ontologies and databases, including:

\begin{itemize}
    \item UMLS: Unified Medical Language System CUIs for both diseases and drugs (in total 8,542 drug entities and 119,458 disease entities; 14,850 treatment relationships between these two types of entities);
    \item ATC: Anatomical Therapeutic Chemical codes for drug classification (6,440 entities in total); 
    \item ICD9-CM: Diagnostic codes for diseases (22,410 entities in total);
    \item MONDO: Harmonized disease ontology (28,090 entities in total);
    \item DrugBank: Chemical and pharmacological drug information (16,581 entities in total);
    \item DrugCentral~\cite{ursu2016drugcentral}, RepoDB~\cite{brown2017standard}, PrimeKG~\cite{chandak2023building}: Sources of curated disease-drug indications. DrugCentral contains 11,138 relationships between Drugbank and UMLS Disease CUIs; RepoDB contains 13,558 relationships between Drugbank and UMLS Disease CUIs; PrimeKG contains 7,659 treatment relationships between Drugbank and MONDO
\end{itemize}

From these sources, we normalized the treatment relationship between different types of entities with a multi-hop search combining code matching and treatment query in the Neo4j graph database. Finally, we get a common treatment relationship format based on ICD9-CM and ATC codes, which serves as the ground truth reference KG for downstream comparison. In total, we have 105,290 drug-disease treatment relationships mapping ATC to ICD codes in the reference KG. We select ATC and ICD since they are widely-adopted code systems used in health systems worldwide. 

\paragraph{LLM Querying and Prompt Design}

To explore the capabilities of LLMs in generating disease-treatment associations, we selected several high-performance LLMs: GPT-4o, OpenAI o3-mini, Perplexity-R1, and Perplexity-Sonar. Each model was queried using a standardized set of diagnostic codes (ICD9-CM) as input, and multiple prompt templates were designed to elicit treatment responses in varying formats.

We designed three representative prompt types to examine trade-offs between simplicity, specificity, and structured output:

\begin{itemize}
    \item Prompt A: Given this [\texttt{ICD\_CODE}], please give me the ATC code for the respective disease in a python list.
    
    \item Prompt B:  [\texttt{ICD\_CODE}] give me all the ATC codes at level 5 for the given disease in a python list.
   
    \item Prompt C: For disease \texttt{[DESCRIPTION}] ([\texttt{ICD\_CODE}]), provide treatments in the following JSON format: [\{``Drug Name": ``...", ``ATC Code": ``..."\}, ...]
\end{itemize}

Prompt A requested ATC codes in a simple Python list format, given only an ICD code. Prompt B narrowed the scope by explicitly asking for ATC codes at level 5. This level of specificity is worth evaluating, as ATC level 5 corresponds to the actual chemical substances prescribed and, therefore, aligns with how medications are documented and managed in prescribing systems. Prompt C introduced a more structured instruction, requesting treatments with both drug names and ATC codes in a specific JSON format using the disease description and ICD code. 

\paragraph{Evaluation Types and Metrics}
To assess how well LLM-generated treatment sets align with curated KG, we employ three types of evaluation including coverage, alignment, and robustness check. We leverage different metrics, including success rate, recall, and set-based similarity metrics to conduct evaluation:

\begin{itemize}

  \item \textbf{Success Rate} (coverage check): 1 if the LLM outputs contain at least 1 drug in the KG reference list. 
  \item \textbf{Recall} (coverage check): Measures the proportion of known treatments recovered by the LLM output.
  \[
  \text{Recall} = \frac{|G \cap K|}{|K|}
  \]
  where $G$ is the drug set of treatments generated by the LLM, and $K$ is the corresponding treatment drug set from the knowledge graph.

  \item \textbf{Jaccard Similarity} (alignment and robustness check): Quantifies the overlap between generated and reference sets relative to their union.
  \[
  \text{Jaccard}(G, K) = \frac{|G \cap K|}{|G \cup K|}
  \]

  \item \textbf{Sørensen--Dice Coefficient} (alignment and robustness check): Gives greater weight to shared elements in smaller sets.
  \[
  \text{Dice}(G, K) = \frac{2|G \cap K|}{|G| + |K|}
  \]
\end{itemize}

The rationale for using multiple similarity metrics is to consider their complementary perspectives on set overlap. For the coverage check, we use success rate and recall, two metrics evaluated against the KG reference list. Although the KG is not complete, it contains a valid subset of all potential treatment relationships and thus provides a reasonable reference for evaluating the LLMs' output. The success rate metric provides a coarse-grained, binary assessment of whether the LLM output contains any relevant treatment from the knowledge graph. Success rate is a simple but informative check for minimum task success and is especially helpful in cases where recall might be low but coverage still exists. Recall is particularly suitable for measuring how well LLMs can recover the known treatment relationships from the knowledge graph, offering insight into coverage and clinical completeness. However, it does not penalize the generation of irrelevant or hallucinated items. Jaccard similarity, in contrast, balances coverage and precision by considering both false positives and false negatives through the union of predicted and true sets. This makes it sensitive to overgeneration, which is common in LLM outputs. Sørensen--Dice coefficient is more tolerant of imbalanced set sizes and places greater emphasis on the size of the intersection relative to the average size of the sets. This is particularly useful when either the LLM or the KG returns a small number of treatments. These two similarity metrics are also applied for the robustness check within each LLM's own responses. 

By using all four metrics in parallel, we aim to capture a comprehensive evaluation of LLM alignment with medical knowledge graphs that consider coverage, overgeneration, and robustness across diseases with both sparse and dense treatment mappings. We report the average scores for 30 sampled ICD9CM disease codes weighted by their frequency in an EHR system from a Portugal hospital containing 9,798,305 patient records.

\section{Results}
\label{sec:results}

To have a thorough evaluation, we perform LLM experiments with difference prompts on the completion of the disease-drug treatment relationship in medical knowledge graphs through three dimensions: coverage, alignment, and robustness. These aspects assess the completeness of the generated relationships, the consistency with existing clinical knowledge, and the stability under various interactions.

\begin{figure}[H]
  \centering
  \begin{subfigure}[b]{0.31\textwidth}
  \includegraphics[width=\textwidth]{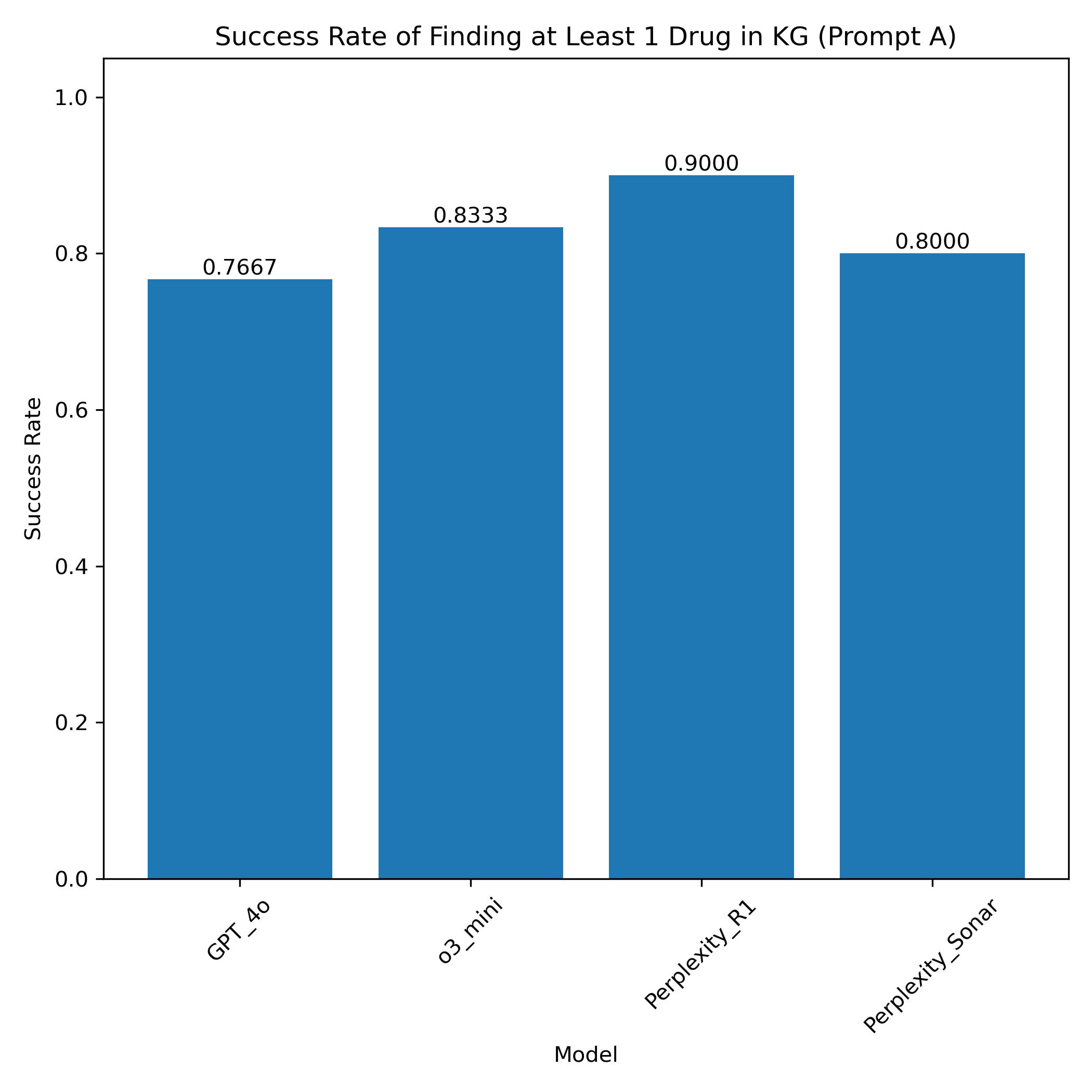}
    \caption{Success Rate (Prompt A)}
    \label{fig:succA}
  \end{subfigure}
  \hfill
  \begin{subfigure}[b]{0.31\textwidth}
  \includegraphics[width=\textwidth]{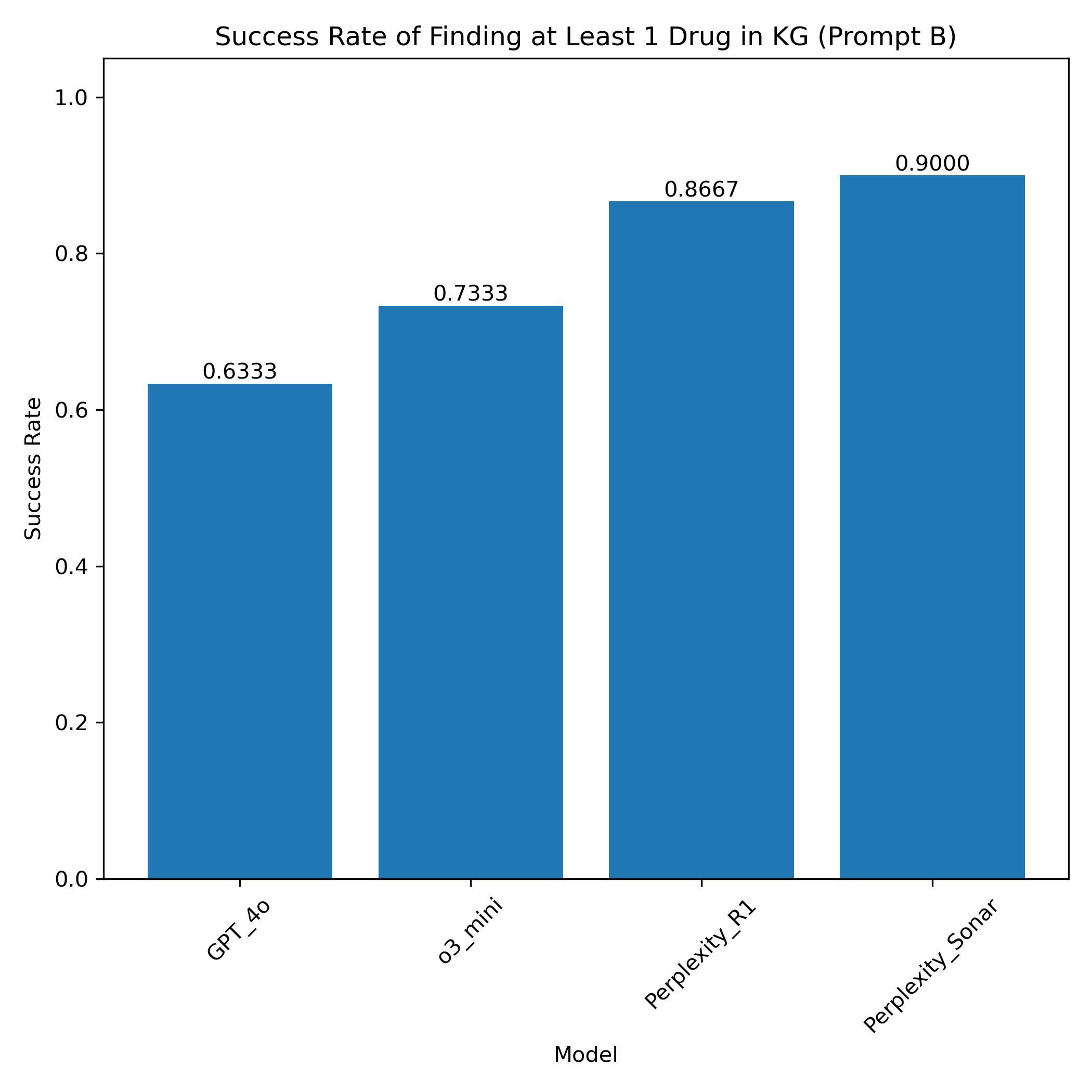}
    \caption{Success Rate (Prompt B)}
    \label{fig:succB}
  \end{subfigure}
  \hfill
  \begin{subfigure}[b]{0.31\textwidth}
  \includegraphics[width=\textwidth]{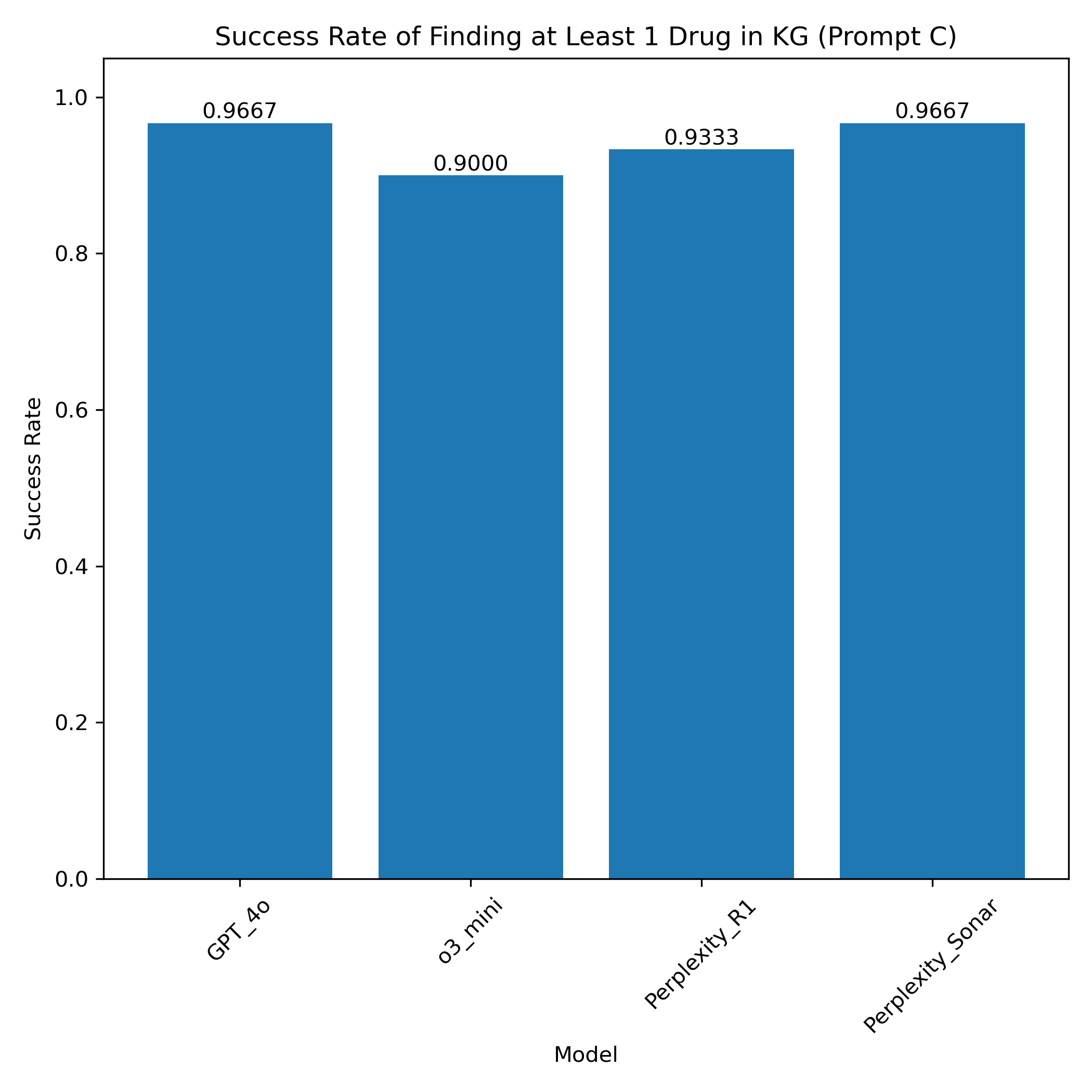}
    \caption{Success Rate (Prompt C)}
    \label{fig:succC}
  \end{subfigure}
  \vfill
  \vspace{2em}
  \begin{subfigure}[b]{0.31\textwidth}
  \includegraphics[width=\textwidth]{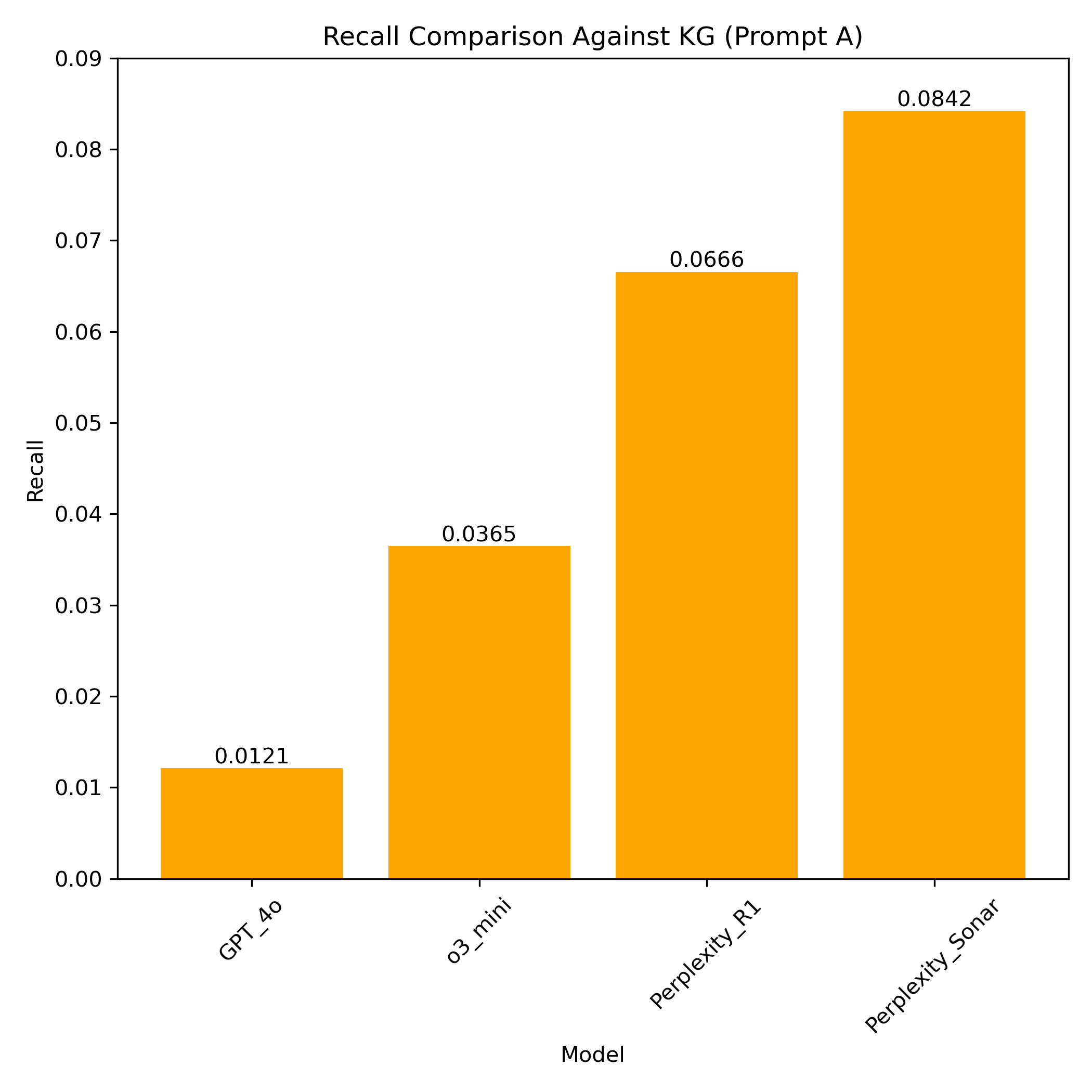}
    \caption{Recall (Prompt A)}
    \label{fig:recallA}
  \end{subfigure}
  \hfill
  \begin{subfigure}[b]{0.31\textwidth}
  \includegraphics[width=\textwidth]{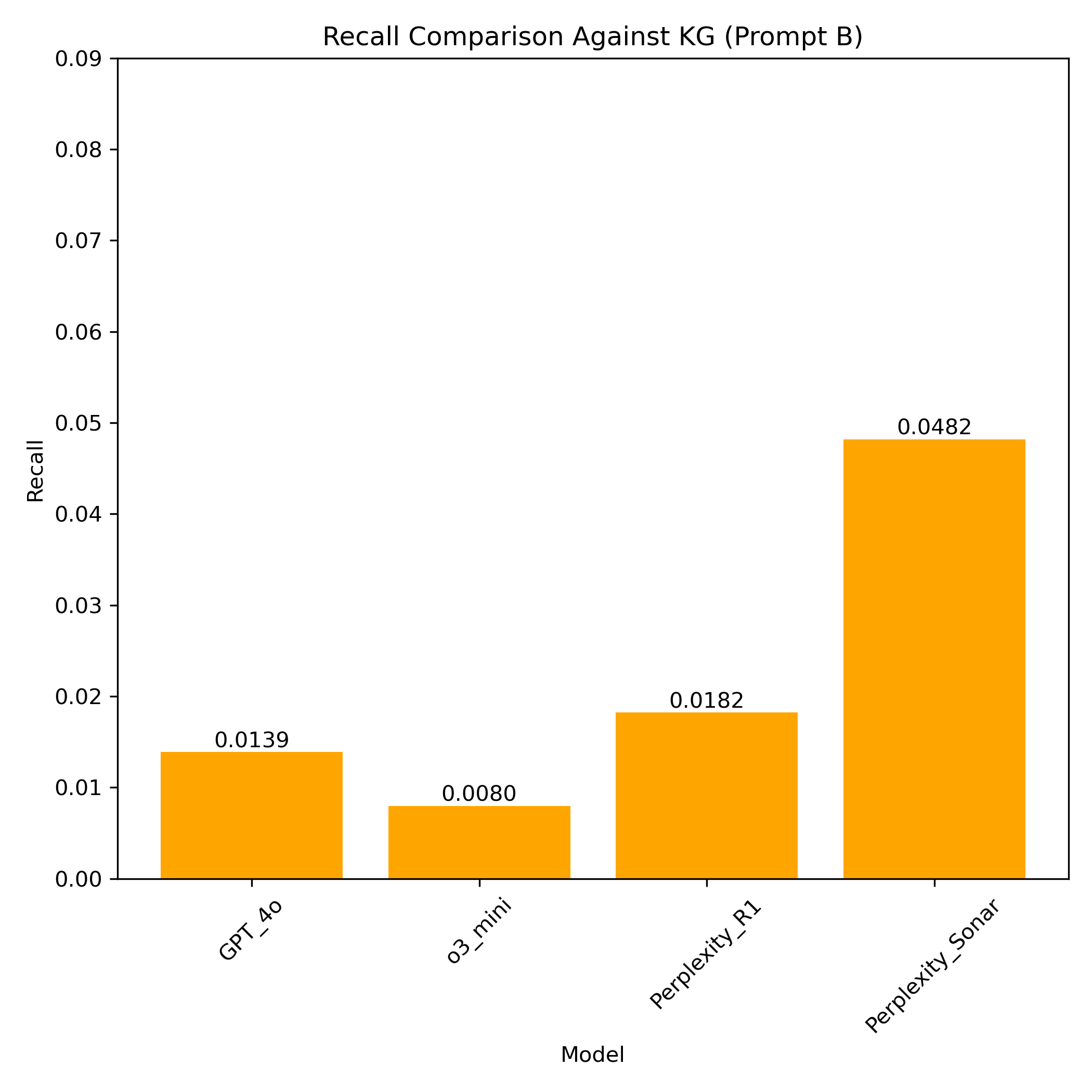}
    \caption{Recall (Prompt B)}
    \label{fig:recallB}
  \end{subfigure}
  \hfill
  \begin{subfigure}[b]{0.31\textwidth}
  \includegraphics[width=\textwidth]{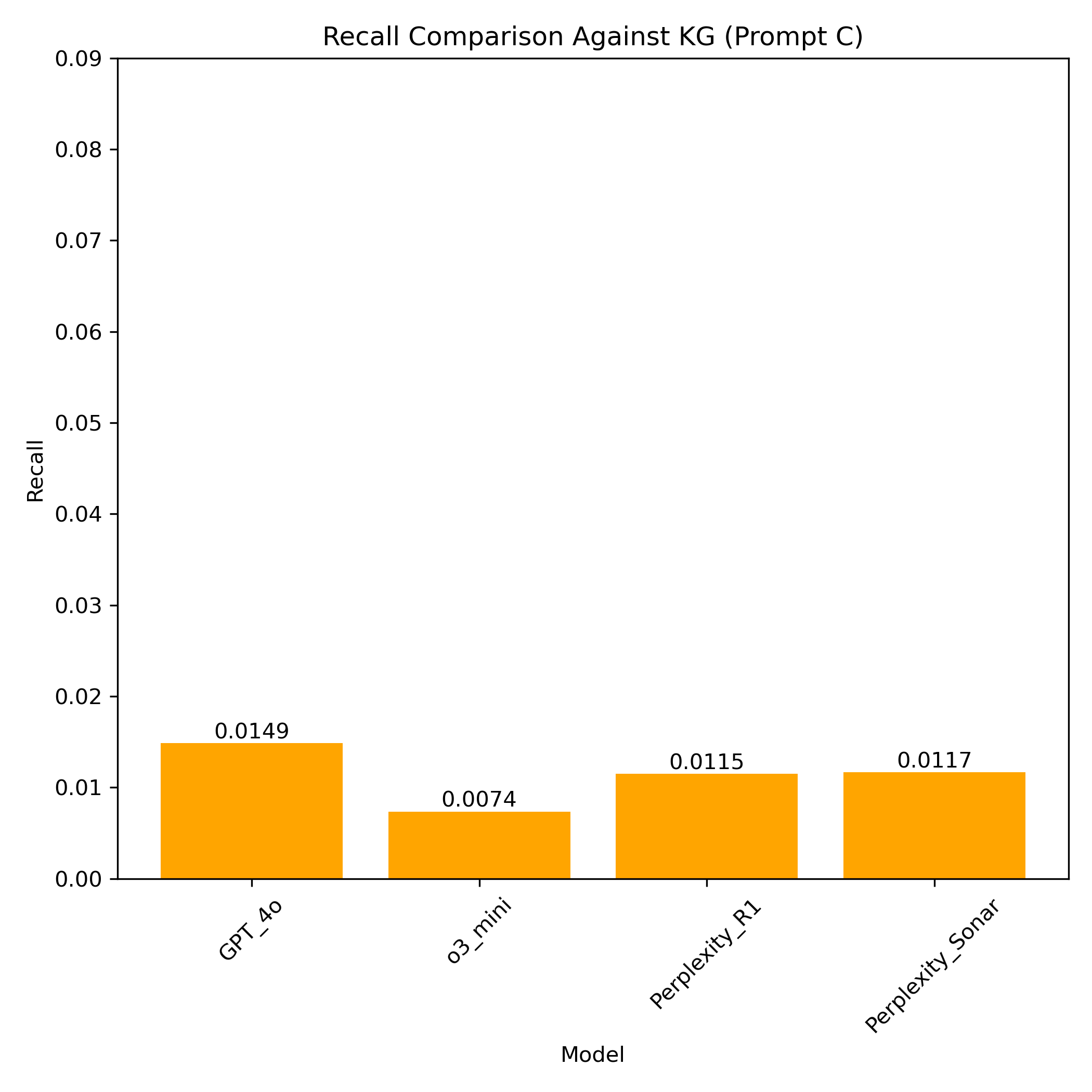}
    \caption{Recall (Prompt C)}
    \label{fig:recallC}
  \end{subfigure}
  \vspace{1em}
\caption{Coverage Performance of LLMs against KG Reference List for Three Prompts}
  \label{fig:coverage_results}
\end{figure}

\paragraph{Coverage Performance of LLMs against KG reference} 
Figure~\ref{fig:coverage_results}  illustrates the average performance of GPT-4o, o3 mini, Perplexity-R1, and Perplexity-Sonar, accessed via API calls, across three prompt formulations (A, B, and C). The evaluation is based on the success rate and recall, measured against the reference KG introduced described in Section~\ref{sec:dm}. 

For the success rate, Prompt C consistently yields the highest success rate in all models evaluated among the three prompts, shown in Figure~\ref{fig:succA}-\ref{fig:succC}. Notably, in Prompt C, Perplexity Sonar and GPT-4o achieve a high success rate of 0.9667, while the remaining models also exceed 0.90, indicating that Prompt C is particularly effective at eliciting partially correct clinical content with more requirements on a structured output format. Prompts A and B achieve relatively worse success rates. For instance, while Perplexity-R1 on Prompt A and Perplexity Sonar on Prompt B achieve 0.90, GPT-4o only achieves 0.7667 and 0.6333 on these two prompts, showing a large variability. Also, in most cases, the success rate of Prompt B is lower than that of Prompt A, which might indicate that detailed instruction as ``level 5" limited the LLMs to give a wider output range.   

A more stringent evaluation using recall reveals a substantial performance gap, shown in Figure~\ref{fig:recallA}-\ref{fig:recallC}. Despite relatively high success rates, all models show low recall, highlighting their limited ability to comprehensively identify the expected set of treatment options. Prompt A yields the highest overall recall, with Perplexity Sonar achieving 0.0842, while other models fall below 0.07. Prompt B, despite requesting more specific ATC level 5 codes, results in even lower recall values in most cases (e.g., o3-mini: 0.0365 $\rightarrow$ 0.0080; Preplexity-R1: 0.0666$\rightarrow$ 0.0182). Prompt C, although the most effective in achieving partial correctness, records the lowest recall scores, with model performance ranging from 0.0074 to 0.0149. This suggests that its structured format may prompt models to generate a small number of well-formed but limited outputs, prioritizing format adherence over comprehensive retrieval.

These findings underscore a critical limitation of current LLMs in clinical decision-making contexts. Although high success rates suggest that models are often capable of retrieving at least one correct treatment option, their low recall indicates a failure to comprehensively capture the full spectrum of clinically appropriate drugs. Furthermore, results suggest that increased specificity, such as prompting for ATC level 5 drugs, does not yield consistent improvements and may restrict model responses in some cases. This indicates that the ability of LLMs to identify partially correct treatment candidates is not necessarily enhanced by more detailed prompting and that general descriptions, as seen in Prompt A, may provide broader cues for model retrieval. This discrepancy highlights the need for approaches that go beyond prompt engineering and address the underlying limitations in domain-specific knowledge retrieval and grounding. 

\paragraph{Alignment Performance Across LLMs} 

\begin{figure}[!htpb]
    \centering
  \begin{subfigure}[b]{0.45\textwidth}
    \includegraphics[width=\textwidth]{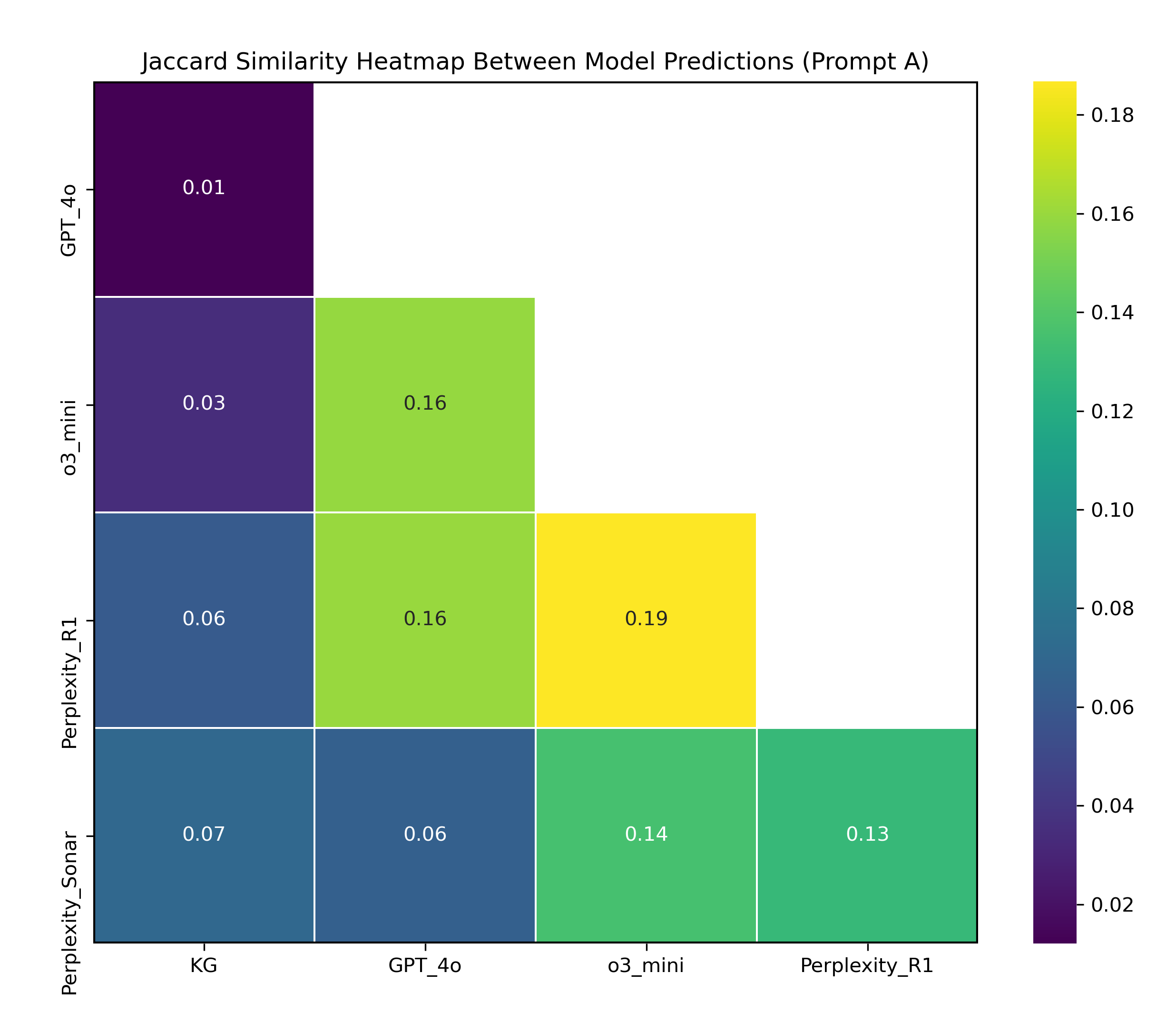}
    \caption{Jaccard Similarity (Prompt A)}
    \label{fig:jaccardA}
  \end{subfigure}
  \hfill
  \begin{subfigure}[b]{0.45\textwidth}
    \includegraphics[width=\textwidth]{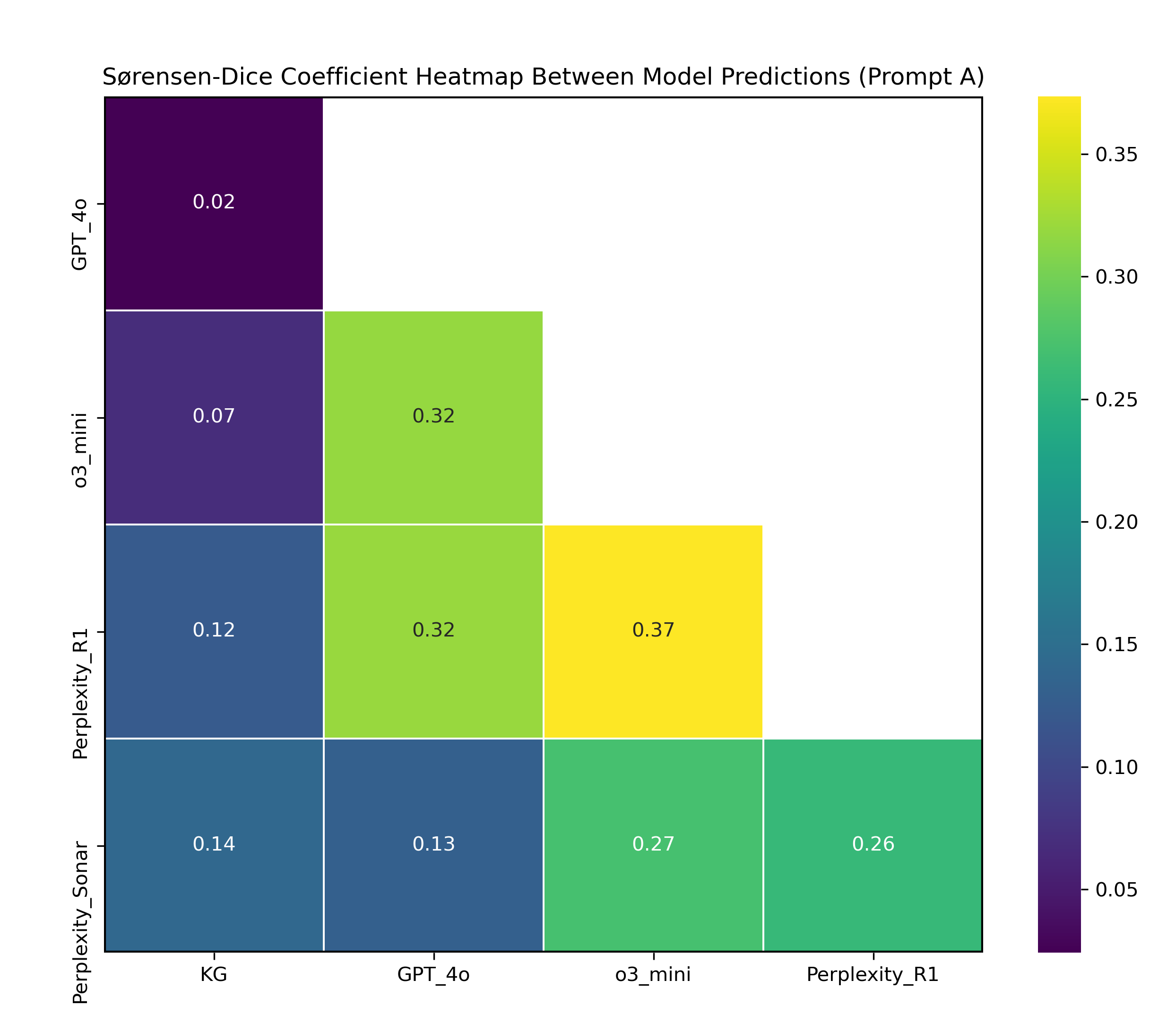}
    \caption{Sørensen-Dice Coefficient (Prompt A)}
    \label{fig:diceA}
  \end{subfigure}
  \vfill
  \vspace{2em}
  \begin{subfigure}[b]{0.45\textwidth}
    \includegraphics[width=\textwidth]{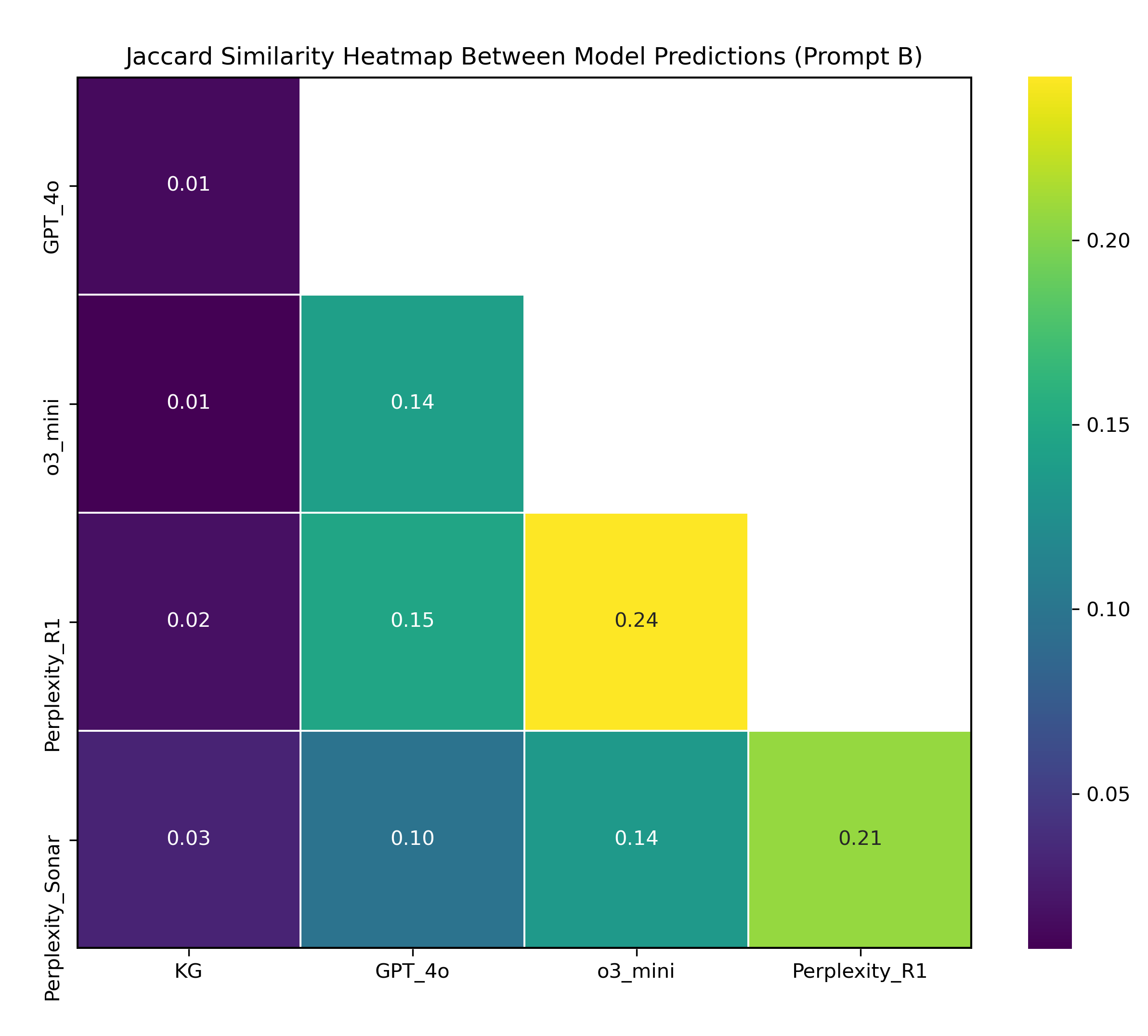}
    \caption{Jaccard Similarity (Prompt B)}
    \label{fig:jaccardB}
  \end{subfigure}
  \hfill
  \begin{subfigure}[b]{0.45\textwidth}
    \includegraphics[width=\textwidth]{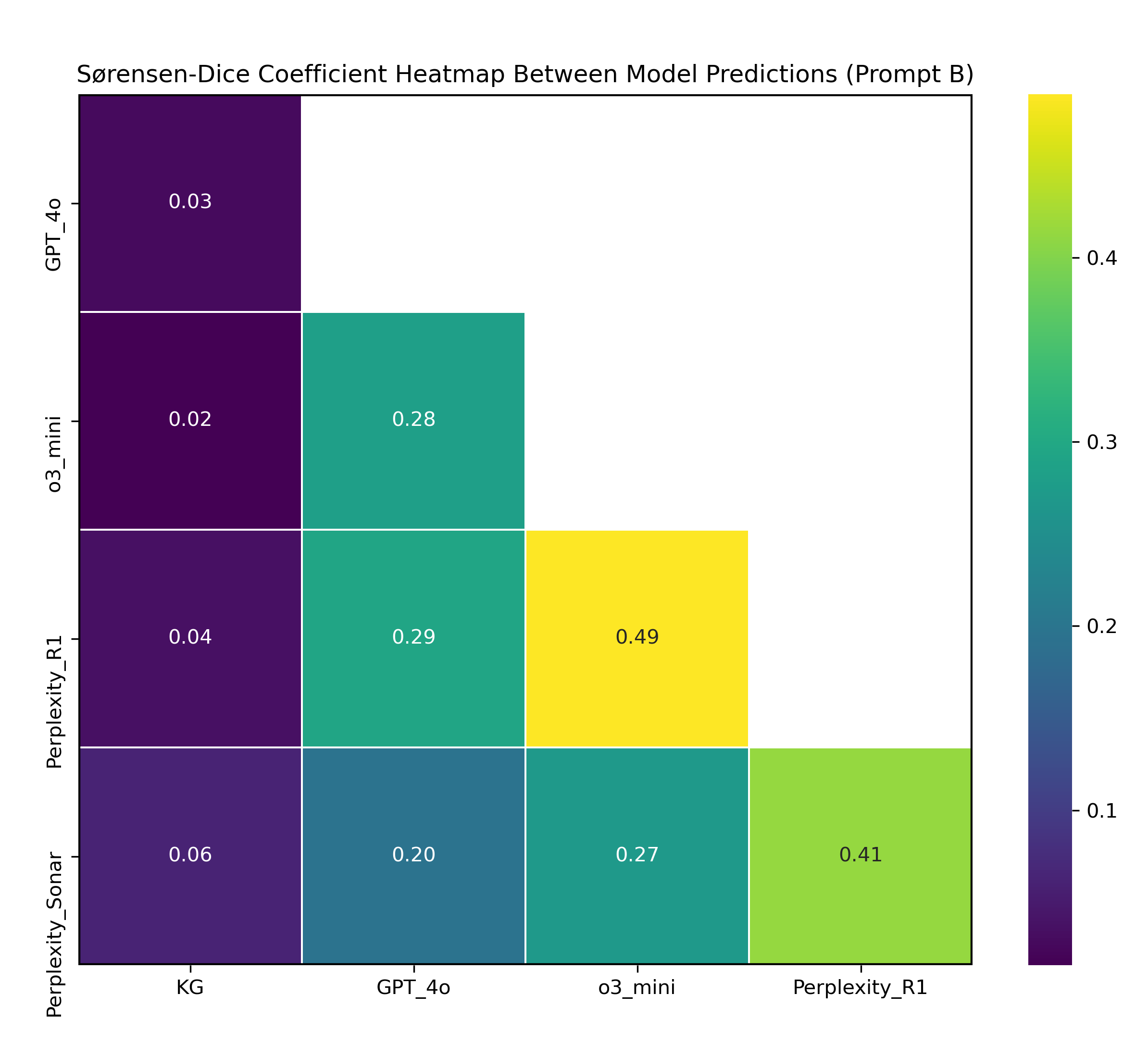}
    \caption{Sørensen-Dice Coefficient (Prompt B)}
    \label{fig:diceB}
  \end{subfigure}
  \vfill
  \vspace{2em}
  \begin{subfigure}[b]{0.45\textwidth}
    \includegraphics[width=\textwidth]{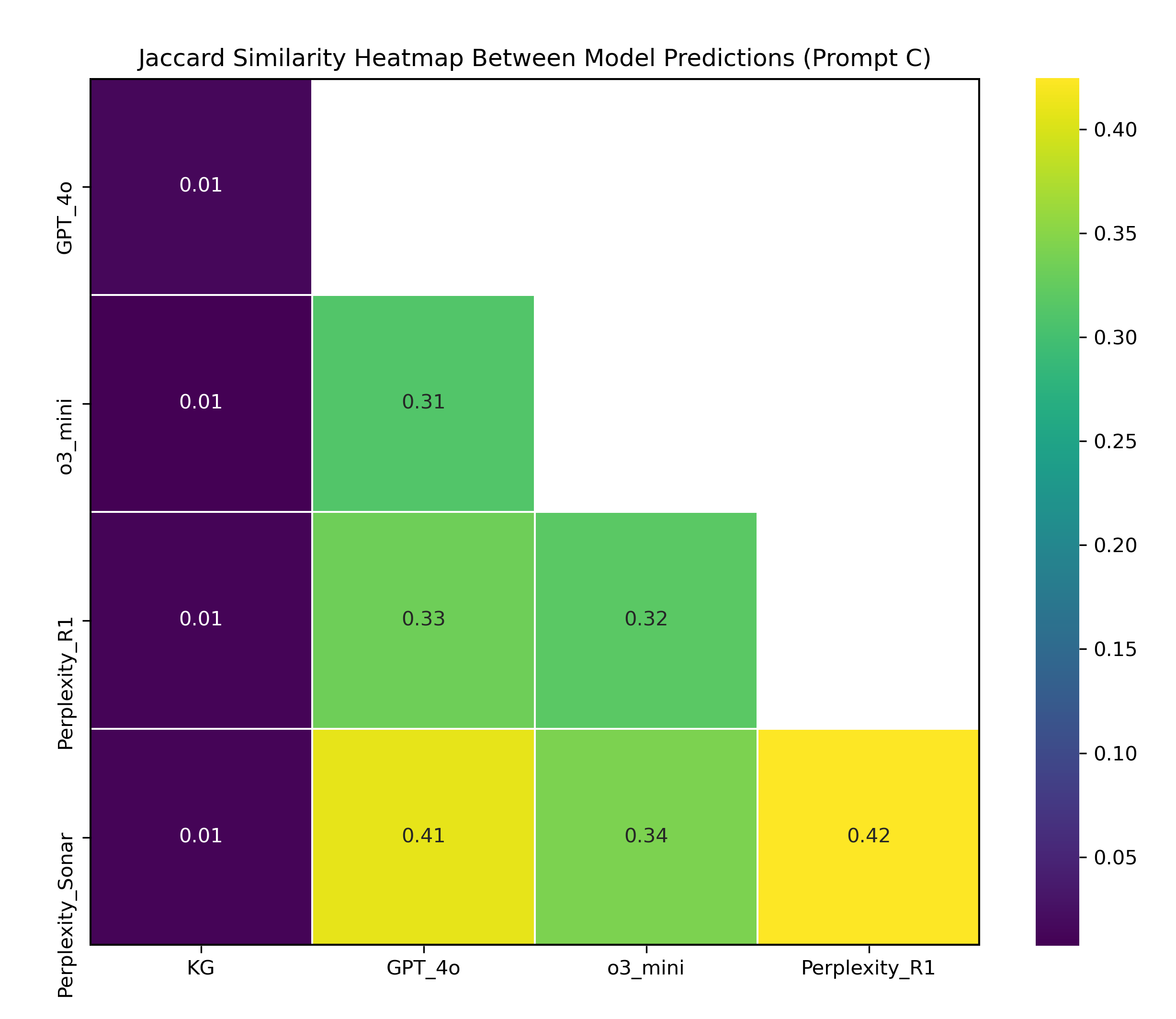}
    \caption{Jaccard Similarity (Prompt C)}
    \label{fig:jaccardC}
  \end{subfigure}
  \hfill
  \begin{subfigure}[b]{0.45\textwidth}
    \includegraphics[width=\textwidth]{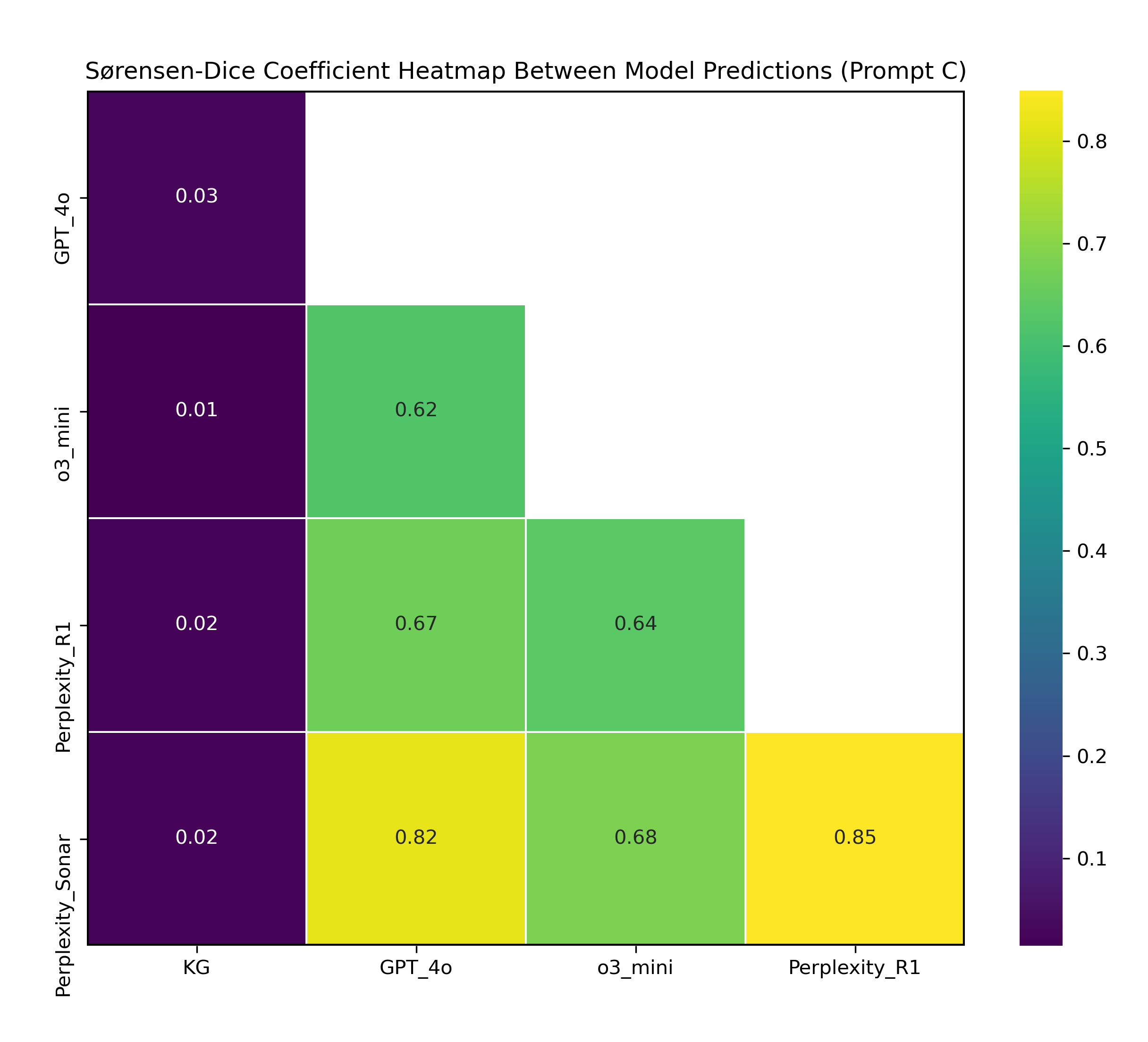}
    \caption{Sørensen-Dice Coefficient (Prompt C)}
    \label{fig:diceC}
  \end{subfigure}
  \vspace{1em}
    \caption{Alignment Performance of Accross LLMs and KG Reference List for Three Prompts}
    \label{fig:alignment_result}
\end{figure}

Figure~\ref{fig:alignment_result} presents the average alignment performance in four LLMs as well as the KG reference list for three prompt formulations evaluated in terms of Jaccard similarity and Sørensen-Dice Coefficient. 

The results of Jaccard similarity and Sørensen–Dice coefficients reveal two key patterns: increased consistency among LLMs with more detailed prompts, accompanied by a decline in alignment with the reference KG list. Under Prompt A, although the inter-LLM agreement is relatively low, with the highest Dice score of 0.32 (between Perplexity-R1 and o3-mini), the alignment with KG is comparatively stronger than that observed under Prompts B and C. Specifically, the highest Dice similarity with the KG is achieved by Perplexity Sonar at 0.14, followed by Perplexity-R1 at 0.12, suggesting that Prompt A elicits predictions that are more aligned with the reference standard. In contrast, Prompt B, with more specific instruction, yields moderately improved consistency among models, as indicated by a Dice coefficient of 0.49 between Perplexity-R1 and o3-mini. However, their alignment with the KG drops (e.g., Perplexity-Sonar's Dice score with KG decreases to 0.06, Perplexity-R1 drops to 0.04). Finally, Prompt C, which contains the most detailed instructions, results in the highest model-to-model similarity, with Dice scores as high as 0.85 between Perplexity-R1 and Perplexity-Sonar and 0.82 between GPT-4o and Perplexity-Sonar. However, it shows the poorest alignment with the KG, with Dice scores below 0.03 for all models. The Jaccard similarity trends mirror those of the Dice coefficient across all prompts. These findings suggest that while more explicit prompts improve the coherence and agreement among LLMs, they may inadvertently reinforce shared reasoning patterns or biases that diverge from the ground truth encoded in the KG. Thus, greater instruction does not guarantee better factual accuracy, highlighting a trade-off between model consistency and reference alignment.

\begin{figure}[!ht]
  \centering
  \begin{subfigure}[b]{0.47\textwidth}
     \centering
    \includegraphics[width=\textwidth]{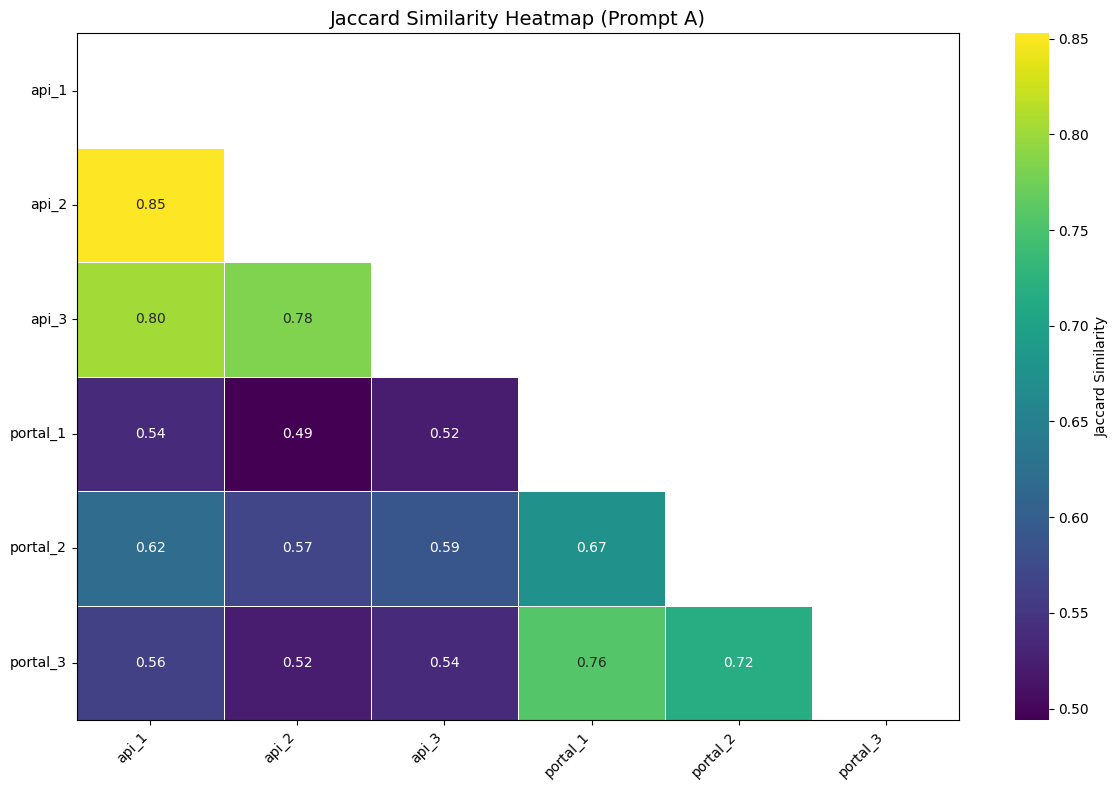}
    \caption{Jaccard Similarity (Prompt A)}
    \label{fig:jaccard_A_robust}
  \end{subfigure}
  \hspace{2em}
  \begin{subfigure}[b]{0.47\textwidth}
    \centering
    \includegraphics[width=\textwidth]{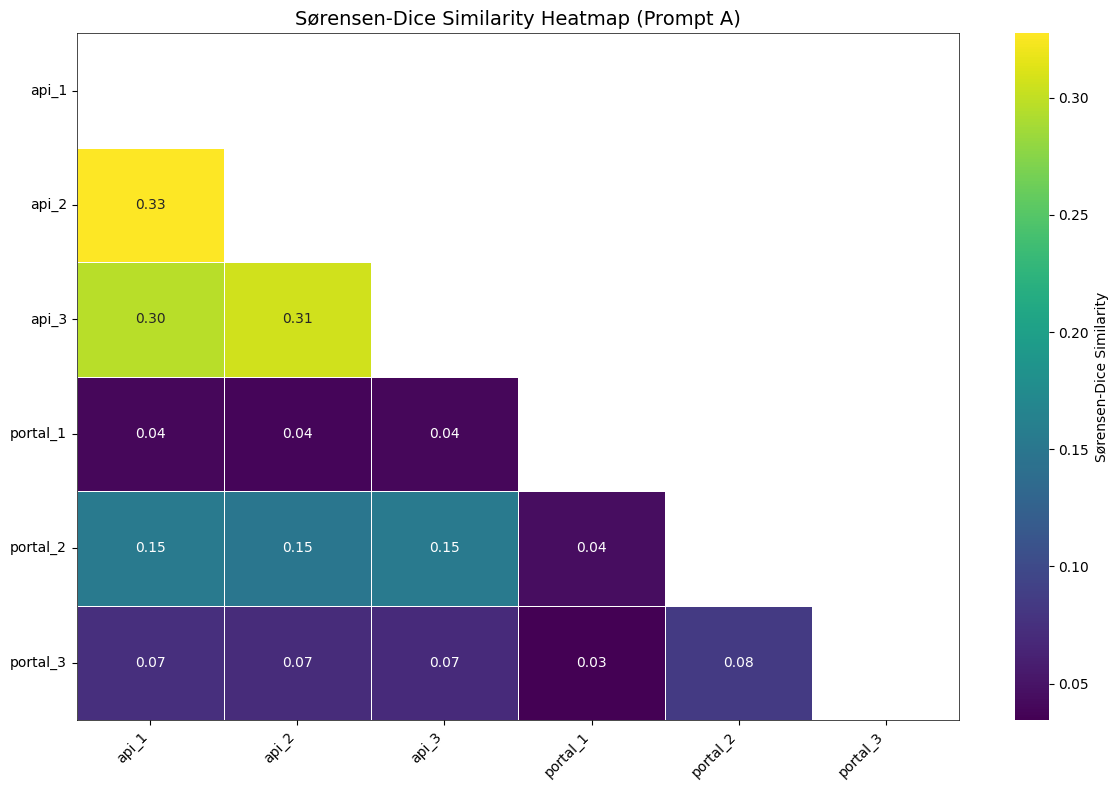}
    \caption{Sørensen-Dice Coefficient (Prompt A)}
    \label{fig:dice_A_robust}
  \end{subfigure}
  \vfill
  \vspace{2em}
   \begin{subfigure}[b]{0.47\textwidth}
   \centering
    \includegraphics[width=\textwidth]{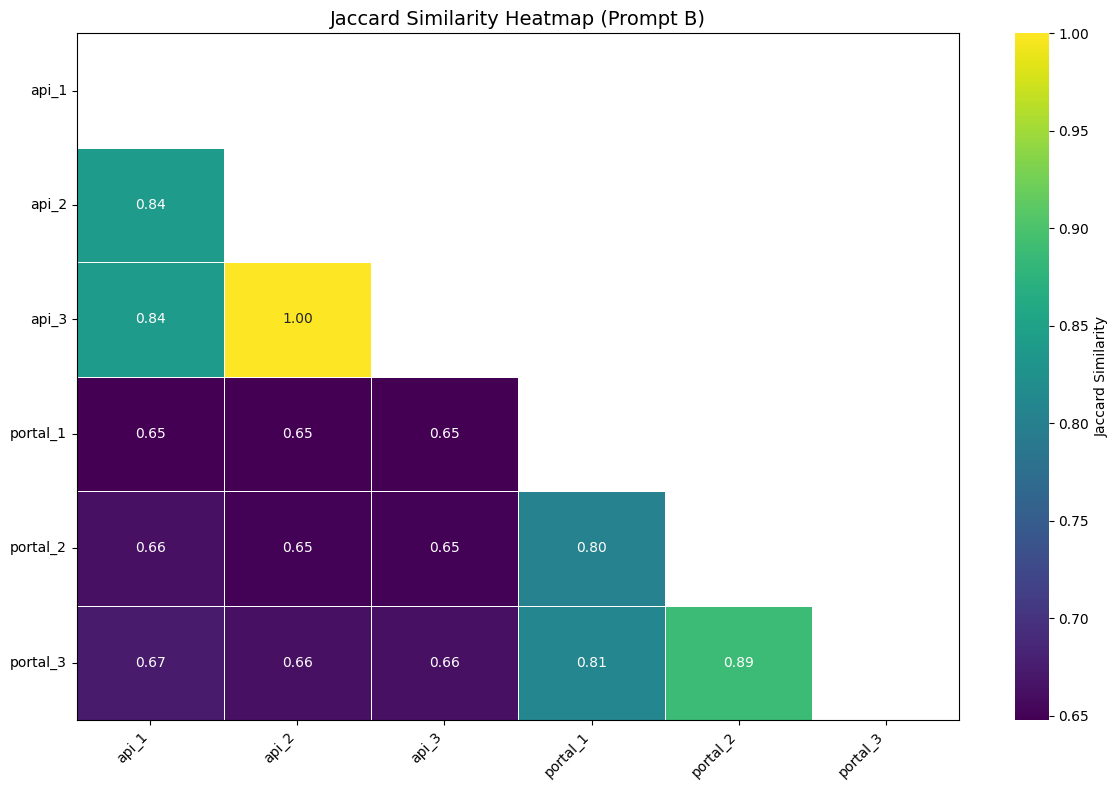}
    \caption{Jaccard Similarity (Prompt B)}
    \label{fig:jaccard_B_robust}
  \end{subfigure}
  \hspace{2em}
  \begin{subfigure}[b]{0.47\textwidth}
  \centering
    \includegraphics[width=\textwidth]{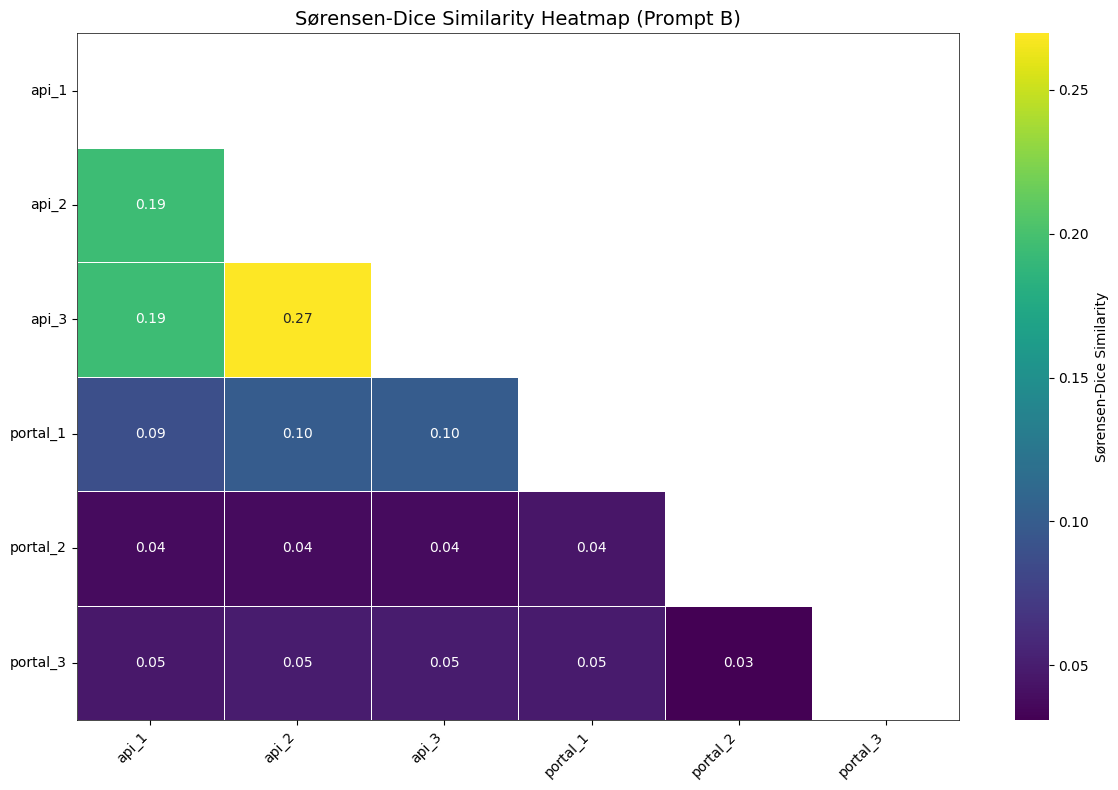}
    \caption{Sørensen-Dice Coefficient (Prompt B)}
    \label{fig:dice_B_robust}
  \end{subfigure}
  \vspace{1em}
  \caption{Consistency and Robustness of Perplexity-Sonar Across Multiple Iterations with API and Web Chat Portal}
  \label{fig:Robustness_result}
\end{figure}

\paragraph{Robustness Check}
In order to comprehensively assess the robustness and consistency of various LLM outputs, we evaluate on prompts A and B, which have relatively high performance in recall and better alignment with the KG reference. For simplicity, we focus on Perplexity-Sonar in this analysis, as it demonstrated comparative superior performance across multiple metrics in earlier sections. We conduct experiments using both the API and the web chat portal provided by the respective vendor. In each setting, we perform three replicated runs to examine whether the results remain consistent or vary over time and across different interaction modes. The results are presented in Figure~\ref{fig:Robustness_result}. 

Notably, despite using the model with the best comprehensive performance, the Sørensen-Dice Coefficient result exhibits inconsistency across different iterations. Our results demonstrate that the API consistently yields relatively uniform responses to a given prompt compared to the web portal since the average pairwise similarity within the API group is higher than that within the web chat portal group for both prompts. Additionally, there is a notable discrepancy between the two groups, as the inter-group similarity is lower than the intra-group similarity for both groups. This disparity can be attributed to the web portal's use of separate threads for each interaction, whereas maintaining a single, continuous thread would have promoted greater consistency.

These findings highlight the challenges of achieving consistent outputs across different platforms, even when using the same LLM vendor. Consequently, the robustness of the LLMs currently, especially for medical data imputation or any disease to drug mapping task are suboptimal.

\section{Conclusion and Future Work}
\label{sec:con}

In this study, we evaluated the ability of LLMs to generate clinically relevant treatment relationships for diseases and assessed how well these outputs align with existing medical KG. By comparing LLM-generated treatment sets with curated KG-derived relationships, we revealed both the potential and the limitations of using LLMs for medical knowledge augmentation. Our evaluation, based on recall and a suite of set-based similarity metrics, demonstrated that while LLMs can recover a substantial portion of known treatments, their performance varies across models and prompt formats. We also observed instances of overgeneration and hallucination, underscoring the importance of rigorous, knowledge-grounded evaluation when deploying LLMs in clinical applications.

The use of multiple similarity metrics allowed us to capture different facets of alignment, including coverage, precision, and robustness to set size imbalance. Our findings suggest that although current LLMs possess impressive generalization capabilities, their outputs must be carefully validated, particularly in high-stakes domains like medicine, where factual correctness is critical.

\paragraph{Future work} We aim to improve KG imputation by exploring retrieval-augmented generation (RAG) as a more controllable and interpretable alternative to open-ended LLM prompting. By focusing the generation on structured knowledge sources such as ontologies, drug databases, and clinical guidelines, RAG has the potential to improve the consistency of facts and reduce hallucinated output. We also plan to investigate techniques for automatic confidence scoring or self-verification of LLM responses, allowing a more selective integration of generated content into medical knowledge graphs. Also, we are interested in developing task-specific evaluation datasets that go beyond treatment mapping to assess the broader applicability of LLMs in clinical knowledge reasoning across contraindications, polypharmacy management, and care pathway optimization.

Ultimately, our work contributes to a growing understanding of how LLMs interact with structured medical knowledge and highlights the need for hybrid approaches that combine generative capabilities with structured validation and domain-aware reasoning.

\bibliographystyle{unsrt}
\bibliography{ref}

\end{document}